\documentclass{article}
\usepackage[preprint]{neurips_2025}
\usepackage[utf8]{inputenc} 
\usepackage[T1]{fontenc}    
\usepackage{hyperref}       
\usepackage{url}            
\usepackage{booktabs}       
\usepackage{amsfonts}       
\usepackage{nicefrac}       
\usepackage{microtype}      
\usepackage{xcolor}         
\usepackage{enumitem}

\usepackage{microtype}
\usepackage{graphicx}
\usepackage{subfigure}
\usepackage{multirow} 

\usepackage{amsmath}
\usepackage{amssymb}
\usepackage{mathtools}
\usepackage{amsthm}
\usepackage[capitalize,noabbrev]{cleveref}
\title{Large Language Models Inference Engines based on Spiking Neural Networks}

\author{%
 Adarsha Balaji\\
 Mathematics and Computer Science Division\\
 Argonne National Laboratory\\
 Lemont, IL 60439 \\
  \texttt{abalaji@anl.gov} \\
\And
  Sandeep Madireddy \\
  Mathematics and Computer Science Division \\
  Argonne National Laboratory\\
  Lemont, IL 60439 \\
  \texttt{smadireddy@anl.gov} \\
\And
  Prasanna Balaprakash \\
  Computer Science and Mathematics Division \\
  Oak Ridge National Laboratory\\
  Oak Ridge, TN, 37830 \\
  \texttt{pbalapra@ornl.gov} \\
}

\newcommand{\tech}{\text{{NeuTransformer}}}{}

\begin{document}

\maketitle

\begin{abstract}
Foundational models based on the transformer architecture are currently the state-of-the-art in general language modeling, as well as in scientific areas such as material science and climate. However, training and deploying these models is computationally challenging as the time and space complexity has a quadratic relation to the input sequence length. Several efforts exploring efficient computational paradigms and model architectures to address these limitations have been made. 

In this work, we explore spiking neural networks (SNNs), an energy-efficient alternative to traditional neural networks, to design transformer models. A challenge in training large-scale SNNs, such as foundational models, using existing surrogate learning methods is inefficient and time-consuming. On the other hand, techniques to convert existing transformer-based models to their SNN equivalent are not scalable, as achieving optimal performance comes at the cost of a large number of spike time-steps, i.e. increased latency. To address this, we propose \tech, a methodology for designing transformer-based SNN for inference using a supervised fine-tuning approach with existing conversion methods. The proposed methodology works in three steps: (1) replacing the self-attention mechanism with a spike-based self-attention (SSA), (2) converting the feed-forward block of the trained transformer model to its equivalent SNN, and (3) fine-tuning the SSA block using SNN-based surrogate learning algorithms. We benchmark the proposed methodology and demonstrate its accuracy and scalability using three variants of the GPT-2 model of increasing model size. We observe that the converted GPT-2 small models demonstrate a 5 - 12\% loss in cosine similarity and a 9.7\% reduction in perplexity. Finally, we demonstrate the energy efficiency of the SSA block compared to the ASA block and show between 64.71\% and 85.28\% reductions in estimated energy consumption when implementing the self-attention mechanism on a digital hardware.

\end{abstract}


\section{Introduction}\label{sec:intro}

Recent advances in Artificial Intelligence (AI) have transformed our approach to solving scientific problems. This is particularly evident in the development of AI foundational models for use in the fields of natural language processing \cite{vaswani2017attention}, material science \cite{boiko2023autonomous}, cancer research \cite{zhou2022cancerbert} and weather/climate \cite{nguyen2023climax, kraus2023enhancing}. These foundational models are traditionally designed using the transformer architecture, a sequence-to-sequence model based on the multi-headed self attention mechanism (SA). However, the transformer is a memory and computation intensive block, leading to the need for expensive and memory constrained AI hardware, such as GPUs or custom accelerators, to train and infer these models. To address the compute demand, we explore spiking neural network (SNNs), a paradigm inspired by the biological concepts of the mammalian brain, to implement self-attention and exploit the data-, energy-, and resource-efficient execution of foundational models on neuromorphic hardware. 

Designing large-scale SNNs often involves two key techniques - (1) convert a pre-trained models to its SNN equivalent \cite{rueckauer2016theory,Diehl2015,cao2015spiking,ho2021tclconversion,midya2019artificial}, where-in, the average firing rate of the SNN neurons are approximated to the activation of the corresponding baseline model's neurons. The converted SNNs can obtain near loss-less accuracy when compared to baseline, but requires an increased number of spike time-steps to reach an accurate estimation thus increasing inference latency, (2) directly train an SNN using back-propagation based learning rules. However, direct training is challenging as computing the SNN's error function is infeasible  due to the discrete nature of its activations (spikes). Several approximate gradient approaches, such as surrogate gradient (SG) \cite{stewart2022meta}, are proposed but are often limited in their ability to learn on large scale (parameters) SNN.

In this work, we aim to exploit both the above proposed methods to implement SNN-based transformers. The key component of the transformer model is the self-attention mechanism. The analog self-attention (ASA) mechanism transforms an input sequence into an attention map. The ASA takes the Query (Q), Key (K) and Value (V) matrices as input and perform three operations on the input: matrix multiplication (dot product), scale and softmax activation. However, the dot product and softmax activation operations cannot be readily implemented in SNNs, due to the binary nature of SNN data (spikes). To address this, a method to train an SNN-based transformer architecture \cite{yao2024spike} is explored and successfully demonstrate spiking self-attention (SSA). However, this method is limited to smaller vision-based transformer models due to the inefficiency of SNN-based back-propagation algorithms when applied to deep learning tasks. To address this, we propose \tech, a method to design spike-based transformer architecture (STA) from trained transformer models followed by the supervised fine-tuning of the attention block of the SNN using surrogate gradient based learning methods. We achieve this in three steps: (1) The ASA block are replaced by the SSA block, (2) converting the feed-forward block of the trained baseline to an SNN, and (3) fine-tuning the SSA block using SNN-based surrogate learning algorithms. The overarching goal of this research is to propose a methodology, \tech, to build data-efficient and energy-efficient SNN-based transformer models to potentially deploy on low-power neuromorphic hardware (NmC).

Following are our key contributions.
\begin{itemize}
    \item A methodology to design transformer-based spiking neural network (SNN) from trained transformer models followed by a supervised fine-tuning of the attention layers of the SNN to improve model performance;
    \item The proposed SNN uses sparse spike-based computation in the self-attention block, replacing the use of energy and latency inefficient matrix multiplication and softmax operations used in the baseline;

    \item By mitigating the need to train the SNN-based transformer model from scratch, we are able to demonstrate novel SNN-based LLMs using the GPT-2 model and its variants of increasing model size. To the best of our knowledge the GPT-2 Large model, designed using \tech, is the largest (parameters) SNN-based transformer model available;

    \item We benchmark the converted SNN model against the baseline model in terms of application accuracy, cosine similarity, perplexity (PPL) and bit-per-byte (BPB) to measure the performance of the model, and energy consumption and throughput to measure the SNN models performance when deployed on a neuromorhpic platform.
\end{itemize}


\section{Related Works}\label{sec:related_works}


Several works have explored training SNN-based transformer architectures to demonstrate the computational efficiency of the model. These works look to expoit the spatio-temporal nature of data represented in SNNs.

In \cite{yao2021temporal}, the authors first suggest a temporal-wise attention module for SNNs to bypass and minimize a few unnecessary input time-steps. The authors then proposed \cite{yao2023attention}, a multi-dimensional attention module along temporal-wise, channel-wise, and spatial-wise separately to optimize membrane potentials, which in turn regulate the spiking response. In \cite{yu2022stsc}, the authors present STSC-SNN, a temporal convolution-based attention mechanisms with an aim to improve spatio-temporal receptive fields of synaptic connections in vision transform models. SCTFA-SNN \cite{wu2023stca} computes channel-wise and spatial-wise attention, separately, to optimize membrane potentials along the temporal dimension. In \cite{yao2023inherent}, the authors propose an advanced spatial attention module to harness SNNs’ redundancy, which can adaptively optimize their membrane potential distribution by a pair of individual spatial attention sub-modules.
Vision transformer-based SNNs, such as Spikformer \cite{li2024spikeformer} proposes a novel spike-based self-attention mechanism called Spiking Self Attention (SSA), using sparse spike-form Query (Q), Key (K), and Value (V) without the use of softmax activation. The SSA is primarily used for vision tasks and used a global average pooling operator to process the vision features input to the encoder block. Spikformer achieves 74.81\% accuracy on ImageNet-1k with four spike time steps, showing the great potential of transformer-based SNNs for the first time. Spikingformer \cite{zhou2023spikingformer} is a modified version of Spikformer with a pre-activation shortcut avoids the multiplications and achieves a lower firing rate. Designed a novel Spike-Driven Self-Attention (SDSA), which used only masks and addition operations without any multiplication, thus significantly reducing the computation energy up to an 87.2-fold decrease compared to the vanilla self-attention. SGLFormer \cite{zhang2024sglformer} proposes an optimized SNN using local and global transformer block to extract features from an input image and a fusion stage to integrate the local and global features extracted. SGLFormer achieves 77.34\% on ImageNet, significantly enhancing the performance of transformer-based SNNs. While these approaches have successfully demonstrated the benefits of spike-based self attention, they are often limited to Vision-based transformer applications. This is due to (1) the increase computational overhead of training large-scale SNN-based models, and (2) due to the limitations of spike-based learning algorithms when applied to deep neural network architectures. 


Addressing the limitation of the above state-of-the-art methods, we design SNN-based transformer models without the need to train the model from scratch by converting pre-trained transformer models into SNNs and fine-tuning them to improve model performance. This reduces the training cost of the SNN model and ensures the scalability of SNN-based transformer models.



  


\section{Background}\label{sec:background}

\begin{figure}[t!]
	\centering
	\centerline{\includegraphics[trim={0 2cm 26.8cm 1cm}, clip, width=0.98\columnwidth]{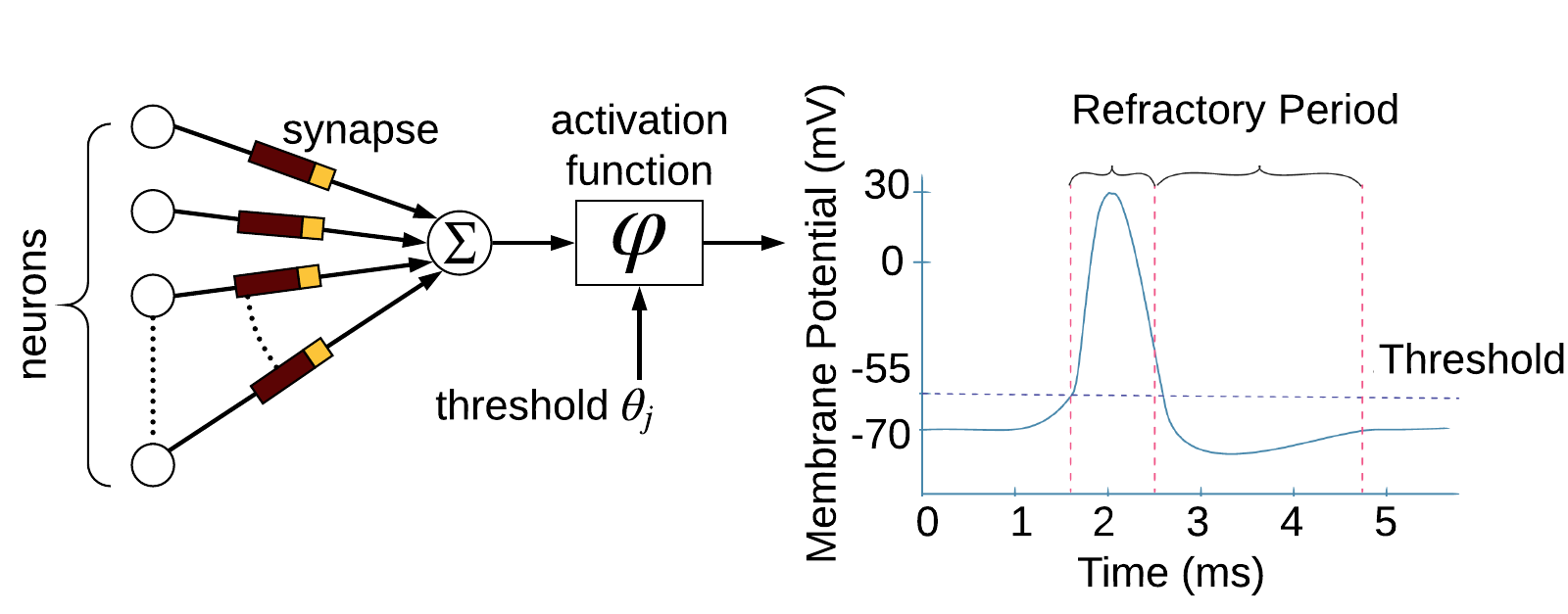}}
	\caption{Overview of a Spiking Neural Network (SNN) and the response function of a spiking neuron.}
	\label{fig:lif}
\end{figure}

In this section, we introduce the concept of spiking neural networks (SNNs), existing ANN-SNN conversion methods and discuss the neuron models used in this work. We also introduce the analog self-attention mechanism and highlight its shortcomings in terms of computational complexity. 

\subsection{Spiking Neural Networks}

Spiking neural networks are event-driven computational models inspired by the mammalian brain. Spiking neurons are typically implemented using variants of the Integrate-and-Fire (I\&F) models \cite{teeter2018generalized} and communicate using spikes. Figure \ref{fig:lif} illustrates an SNN with pre-synaptic neurons connected to a post-synaptic neuron via synaptic elements with weights $w1$, $w2$ respectively. When a pre-synaptic neuron generates a spike, current is injected into the postsynaptic neuron, proportional to the product of the spike voltage and the conductance of the respective synapse. SNNs are trained by adjusting the synaptic weights using a supervised, a semi-supervised, or an unsupervised approach \cite{kasabov2001evolving,mostafa2018supervised}. 

Within the SNN framework, the most representative and widely used neuronal model is the leaky integrate-and-fire (LIF) model. Although the LIF model is only a simplified approximation of real neuronal dynamics, and may not capture all complex neuronal dynamics, its high computational efficiency and spike-response behavior makes it particularly suitable for large-scale neural networks. Key features of the LIF model include the integration of the membrane voltage potential, its inherent leaky nature, and the firing mechanism that activates when a certain threshold is reached. At each time step, the LIF neuron accumulates input currents from previous neurons and changes its membrane potential to represent its active state. When the membrane potential accumulates to a certain threshold, the neuron emits a spike, simulating the firing activities observed in biological counterparts, as shown in equation \ref{eq:lif}. 
\begin{equation}
    \label{eq:lif}
    \tau_m \frac{\partial V_{mem}}{\partial t} = - V_{mem}(t) + R*I(t)
\end{equation}  

where, $V_{mem}$ is the membrane potential, R is the inverse of the weight ($w_i$) of the synapse, and I(t) is the input activation at time-step \textit{t}. The -$V_{mem}$ component is the leaky (non-linear) behavior of the LIF neuron. The LIF neuron can also function as a linear integrator when the leaky component of the LIF neuron is disabled, as shown in equation \ref{eq:if}.

\begin{equation}
    \label{eq:if}
    \tau_m \frac{\partial V_{mem}}{\partial t} = R*I(t)
\end{equation}

This time-controlled behavior not only brings SNNs closer to the authentic operations of biological neural systems, but also makes them more adept than ANNs at learning the spatio-temporal information from event-driven tasks.


\subsection{Attention Mechanism}


The transformer model \cite{vaswani2017attention} is based on the multi-head attention mechanism, comprising several self-attention layers running in parallel. The self-attention mechanism (ASA) uses the matrix dot product and softmax activation functions, as shown in equations \ref{eq:attngeneric}.

\begin{equation}
    \label{eq:attngeneric}
     ASA(Q, K, V) = softmax(Q.K_{T})V
\end{equation}

\begin{equation}
    \label{eq:attn}
    ASA(q_{n}, K, V) =   \sum_{m=1}^{M}\frac {exp(q_{n}^{T}.k_{m})}{\sum_{m=1}^{M} exp(q_{n}^{T}.k_{m})} \cdot v_{m}^{T}
\end{equation}

where, for a given set of inputs $X \to R^{n \times d}$ and trainable parameter matrices $W_{q} \in R^{d \times d_{q}}$ , $W_{k} \in R^{d \times d_{k}}$, $W_{v} \in R_{d \times d_{v}}$, we first calculate the query $Q = XW_{q}$, key $K = XW_{k}$, and value $V = XW_{v}$ matrices respectively. The size of the Q and K matrices is $n \times d_{k}$ and the size of the V matrix is $n \times d_{v}$. The softmax dot-product self-attention operation is defined in equation \ref{eq:attn}. 

The key limitation of self-attention mechanism is the intense computational and memory demands of the dot-product and the softmax operations. The quantized (8-bit or 16-bit) fixed precision multiplication operation used in the dot-product operation scales quadratically with (1) an increase in the context length of the input (n), and (2) the increased dimension ($d_{qkv}$) of the Q, K and V matrices, respectively.
In this work, we propose a spiking implementation of the self-attention mechanism that can replace the inefficient dot-product operation with binary operators and fixed precision accumulators, which will increase the computational demand of the self-attention block, leading to reduced energy consumption and increased throughput.



\subsection{Trained model conversion to SNN}
SNN conversion approaches are proposed in literature to convert trained models to SNN to mitigate the need to retrain a spiking model from scratch and address the limitations of existing SNN learning algorithms to train deep spiking neural networks. The conversion approach aims to map the activation (x) of the neurons in each layer of the ANN to the firing rate (s) of the converted SNN. However, the converted SNN requires large time steps to accurately approximate ReLU activation, which causes large inference latency.

\emph{Spiking ReLU:} Several conversion approaches have been proposed in literature primarily for vision tasks. In early works \cite{perez2013mapping, cao2015spiking}, the authors formalize the relation between the response function of an SNN (LIF) neuron with the activations of the rectified linear unit (ReLU) used in the ANN. They report good conversion accuracy but are restricted to having zero bias and only average-pooling layers. In \cite{Diehl2015}, the authors propose an additional weight normalization approach that achieves near loss-less conversion of ANN-SNN for small networks (MNIST).

\emph{Spiking Softmax:} Softmax activation functions are used in the output layer of the CNN. The softmax activation function generates the probability distribution or the likelihood of the output belonging to a particular class. To replicate this behavior in a spiking neuron, an external spike generator, like a Poisson generator, is used to generate spikes based on the weighted sum accumulated by each spiking neuron.




\section{Methodology}\label{sec:method}

\begin{figure*}[t!]
	\centering
	\centerline{\includegraphics[width=5.7in, height=3.5in]{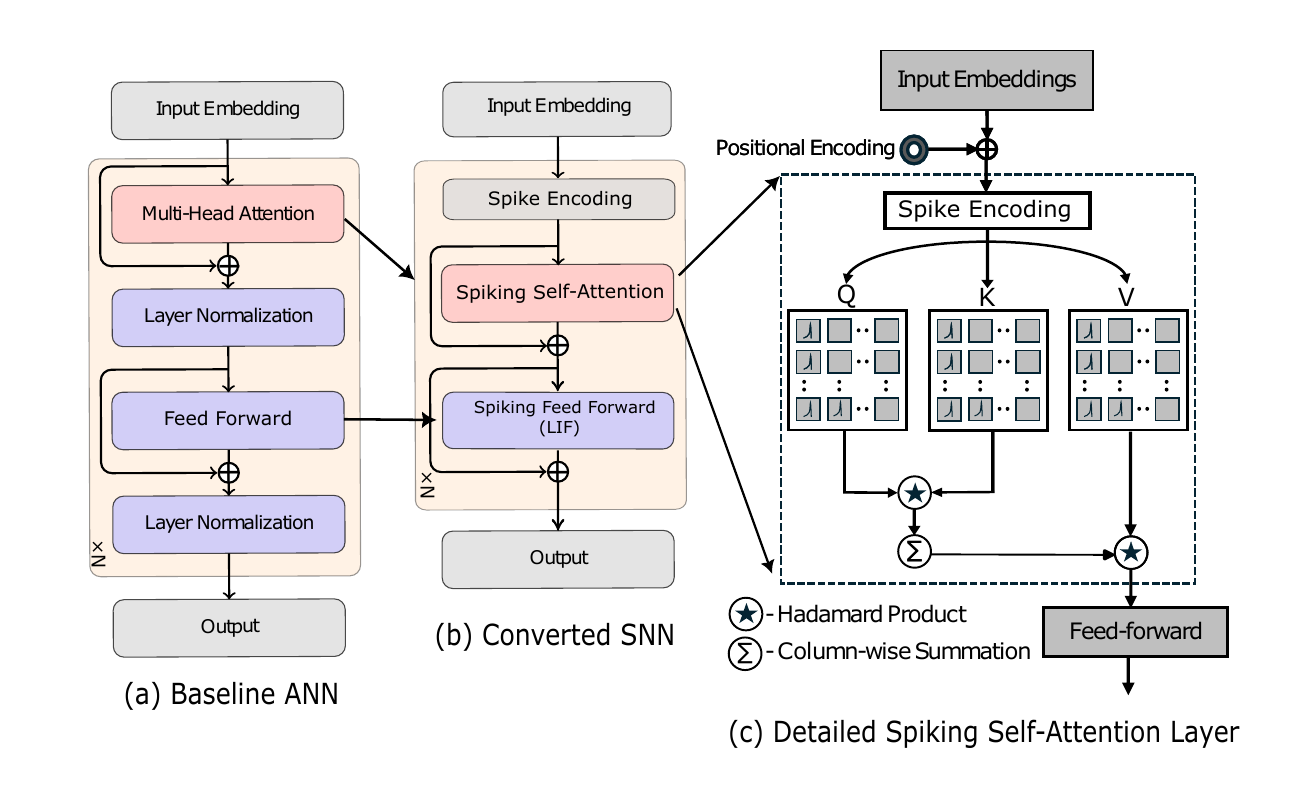}}
	\caption{Conversion of the (a) Baseline Analog Transformer into a (b) SNN Model Architecture. (c) Illustrates the Spiking Self-Attention mechanism, whos operations are fully spiking in nature.}
	\vspace{-5pt}
	\label{fig:conversion_arch}
\end{figure*}

Our approach to convert and fine-tune transformer-based ANN into an SNN consists of \textit{three} key steps: (1) replace the analog self-attention (ASA) block with the proposed spiking-self attention (SSA) block - while retaining the weights of the trained ASA, (2) conversion of the ReLU/GeLU based decision making fully connected layer into an SNN, and (3) fine-tune the SSA block using surrogate learning algorithms. 

\subsection{Spiking Neuron}
The Integrate-and-Fire (IF) neuron model with fixed threshold and adaptable membrane potential decay characteristics is used in this work. In this model, the neuron receives input current, updates its membrane potential, and generates spike output when the membrane potential reaches the threshold. A need of the attention mechanism is for the neuron to process and generate both positive and negative activation (excitatory and inhibitory), and support a fixed threshold in the positive and negative direction. 

\begin{equation}
    \label{eq:neuron}
    \textnormal{S(t)}= 
    \begin{cases}
        1,& \text{if } V_{mem}\geq 1\\
        -1, & \text{if } V_{mem}\leq -1 \\
        0,              & \text{otherwise}
    \end{cases}
\end{equation}

where, $V_{mem}$ is the membrane potential of the neuron and S(t) is the activation (spike) generated at the output of the neuron when the magnitude of $V_{mem}$ crosses the neuron threshold.

\subsection{Spiking Self-Attention}
Figure \ref{fig:conversion_arch} (c) illustrates the fully spiking implementation of the self-attention mechanism. The input to the self-attention block are the spike-encoded sequence (S) generated from the input encodings $I \in {[0,1]}^{T \times N} $, where T is the time window of the input spike train and N is the sequence length of the input. The query (Q), key (K) , and value (V) matrices $\in \mathbb{R}^{T \times N \times D}$, where D ($d_{q,k,v}$) is the model dimension hyperparameter, are computed by performing a linear transformation using three learnable matrices ($W_q$, $W_k$ and $W_v$), respectively, followed by a layer of spiking neurons (LIF) to generate the output spike response. 

Due to the binary nature of the spike encoded activations \textit{I}, the N-bit matrix multiplication in the ASA is replaced by an AND operation followed by an accumulator. The spiking self-attention is computed by performing a column-wise ($d_{k} \times 1$) hadamard operation between the $Q$ and $K^T$ vectors, as shown in equation \ref{eq:SSA1}, followed by a spiking IF activation. The output of the SSA block is computed as the hadamard product of the attention scores generated using equation \ref{eq:SSA1} and the value (V) vector, as shown in equation \ref{eq:SSA2}. In our approach, these weights are initialized with the weights from the trained baseline ANN and tuned to improve the performance of the network. This is discussed in detail in Section \ref{subsec:fine-tune}.


\begin{equation}
    \label{eq:SSA1}
    Attention Score (AS) = LIF((Q \otimes K^T)_{Columnwise})
\end{equation}  

\begin{equation}
    \label{eq:SSA2}
    SSA(Q,K,V) = (AS \otimes V)
\end{equation}


where, Q, K, V are the spike inputs of the Query, Keys and Values and LIF is the spike generator function at the output of the SSA. The generated spikes (SSA) are then passed to the feed-forward layer of the spiking transformer.

\subsection{Spiking feed-forward layer}
The basic principle of converting ANNs into SNNs is that the firing rates of spiking neurons match the graded activations of analog neurons. To achieve this, we start with the relation of ANNs using ReLUs (equation \ref{eq:relu}) and the SNN integrate-and-fire (IF) neuron (equation \ref{eq:lif1}). The ReLU can be considered a firing rate approximation of an IF neuron with no refractory period, whereby the output of the ReLU is proportional to the number of spikes produced by an IF neuron, within a given time window. The first step is to replace the ReLU-based ANN neurons in the feed-forward block with the Integrate-and-Fire (IF) neuron.

\begin{equation}
    \label{eq:relu}
    ReLU(y) = max(0, y)
\end{equation}  

where, y is the weighted sum of the products of the input activation $x_{i}$ and the corresponding weight $w_{i}$.

\begin{equation}
    \label{eq:lif1}
    \tau_m \frac{\partial V_{mem}}{\partial t} = - V_{mem}(t) + R*I(t)
\end{equation}  

Reducing the simulation time-step can help to reduce the number of input spikes per input-step, and increasing the simulation duration will help to avoid insufficient activation. However, all factors can be addressed by finding the right balance of spiking thresholds, input weights and input firing rates.

\emph{Weight Normalization:} Weight normalization is a method used to control the firing rate of SNN neurons. The aim of the weight-normalization process is to optimize the synaptic weights ($W^l$) and the spiking neuron firing threshold ($v_{th}$) such that the firing rate of a spiking neuron is proportional to the activations of its corresponding neuron in the ANN. The normalization is performed layer-wise and the weight normalization factor ($s_{norm}^l$) is set to all the neurons in a layer. In this process, we scale the synaptic weights of the preceding neural layer by a normalization factor $s_{norm}^l$, equal to the maximum positive activation $a_l$ of the ANN neurons. The spiking neuron firing threshold ($v_{th}$) is set as a constant and is equal to 1.
	
To measure the normalization factor, $s_{norm}^l$, for every layer in the neural network we expose the ANN to a batch from the training set. The normalization factor is measured as the  maximum positive activation in each layer.

\begin{equation}
	\label{eq:mapping_rep}
	s_{norm}^l = max(a_l)
\end{equation}
where, l is the corresponding layer in the ANN.  \\
	
The weights are then normalized using
\begin{equation}
	\label{eq:mapping_rep}
	\Tilde{W}^l \longleftarrow \frac{W^l}{s_{norm}^l} 
\end{equation}

\subsection{Fine-Tuning Spiking Self-Attention Block}\label{subsec:fine-tune}
The aim of fine-tuning the model is to minimize the errors induced when we replace the ASA block with the SSA block in step 1. In the ASA block, the output of the softmax operation, as shown in equation \ref{eq:attngeneric}, are the attention scores ($ASA_{as}$) generated using the Q and K matrices. However, the equivalent attention scores generated using the SSA block ($SSA_{as}$), as shown in equation \ref{eq:SSA1}, are the spikes generated by a layer of IF neurons. For a perfectly converted model, we expect the spike rates of the output of the IF neurons to be proportional to the attention scores (softmax scores) generated by the ASA block. However, in practise we observe that the spike rates ($S_r$) deviates from the expected ASA attention score as the functions, equation \ref{eq:attngeneric} and \ref{eq:SSA1} are not identical. Therefore, we propose fine-tuning the weights of the SSA block using a surrogate gradient ($spike_{grad}$) \cite{eshraghian2023training} based learning method with an aim to minimize the loss in accuracy between $ASA_{as}$ and $SSA_{as}$, as shown in equation \ref{eq:loss}.

\begin{equation}
    \label{eq:loss}
    \sum_{i=1}^{d_{model}}(ASA_{as}-SSA_{as})^2
\end{equation}  

where, i is the individual output of the attention score. For this work, we limit the fine-tuning to the SSA block by disabling learning on all other blocks in the model. 



\section{Evaluation Methodology}\label{sec:evaluation}

\begin{figure*}[t!]
	\centering
	\centerline{\includegraphics[width=6in, height=2in]{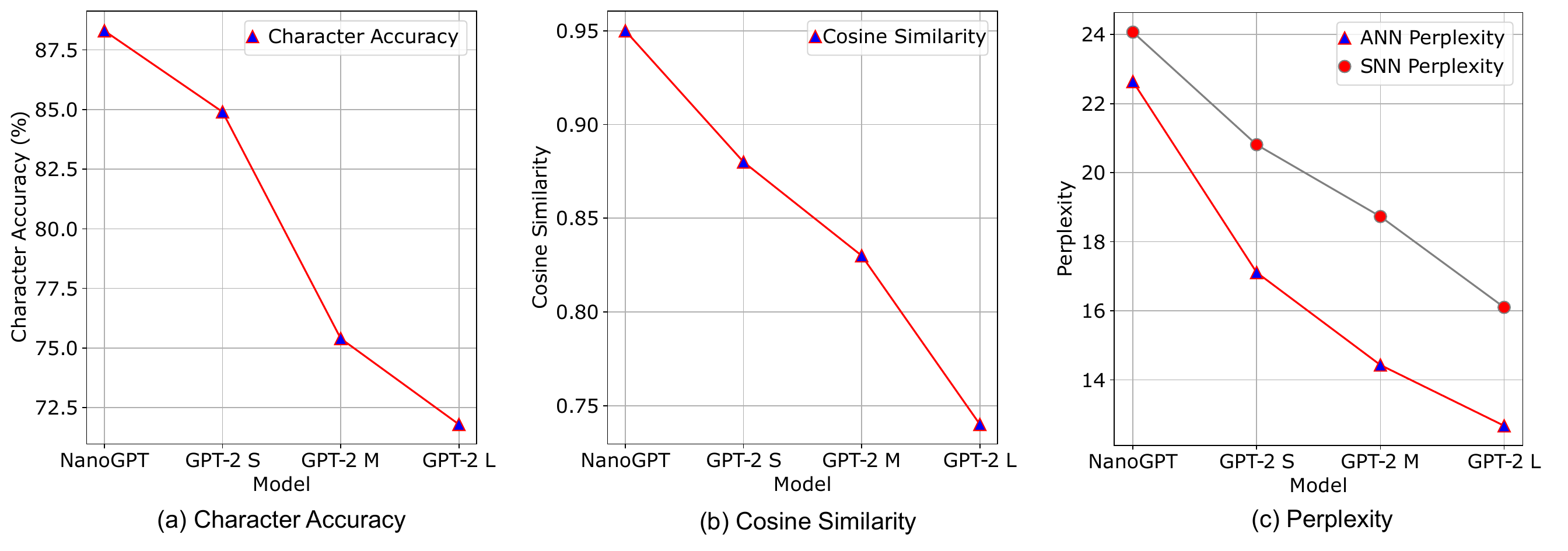}}
	\caption{Performance of the \tech-based implementation of four variants of GPT models - (a) Character Accuracy, (b) Cosine Similarity and (c) Perplexity, when compared to the baseline ANN.}
	\label{fig:plots}
\end{figure*}

\begin{figure*}[t!]
	\centering
	\centerline{\includegraphics[width=2.5in, height=2.3in]{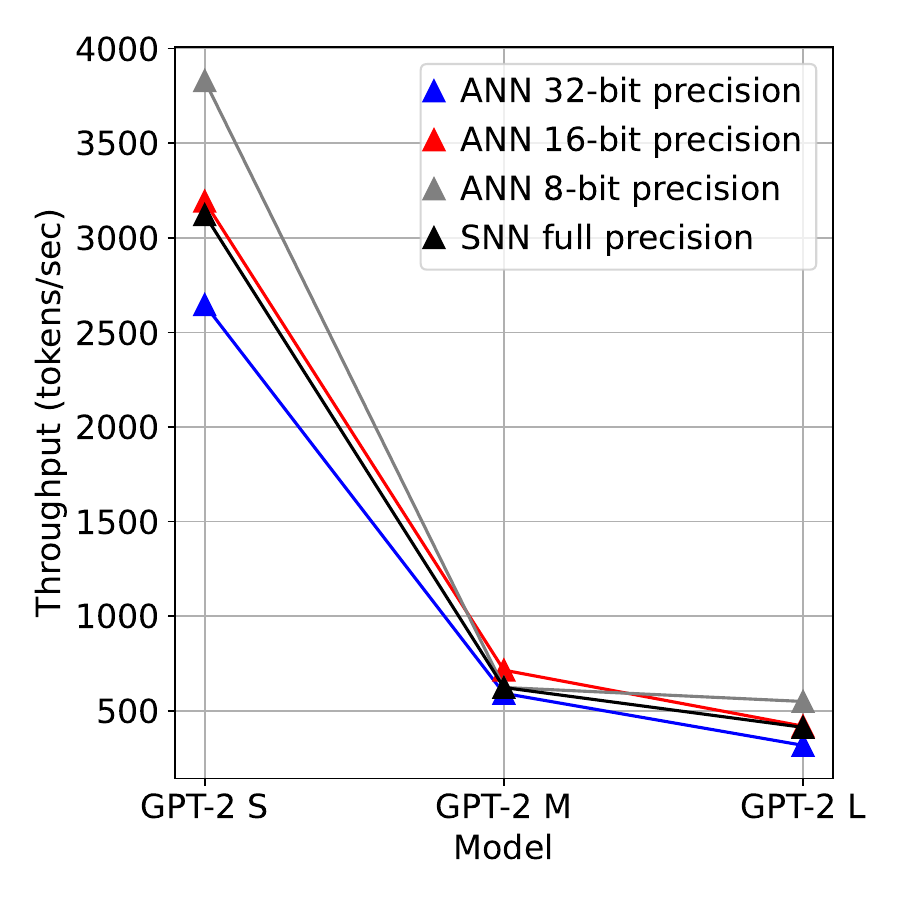}}
	\caption{Throughput of proposed \tech vs baseline ANN models at full-, half- and 8-bit precision.}
	\label{fig:plots}
\end{figure*}

\begin{table*}[t!]
	\renewcommand{\arraystretch}{1.2}
	\setlength{\tabcolsep}{2pt}
	\centering
	{\fontsize{8}{10}\selectfont
	\begin{tabular}{|l||c|c|c|c|c|c|c|c|}
		\hline

		Model & Dataset & Params (M) & Cos Similarity & Char Acc & ANN Perp & SNN Perp & ANN BpB & SNN BpB 
		\\
		\hline
		\hline
		
        GPT-2-Small	&	\multirow{3}{*}{OpenWebText}	&	117	&	0.88	&	84.9	&	17.11	&	21.81	&	2.31	&	2.31 \\
        GPT2-Medium	&		&	345	&	0.83	&	75.4	&	14.43	&	19.73	&	1.97	&	1.97 	\\
        GPT-2-Large	&		&	763	&	0.74	&	71.8	&	12.67	&	18.10	&	1.45	&	1.45 \\
		\hline
	\end{tabular}}
\caption{Comparison of the performance of the SNN w.r.t to baseline transformer model.}
\label{tab:results_conversion}
\end{table*}

\subsection{Evaluation Setup}\label{subsec:setup}
We use the computing resources provided on Swing at the high-performance computing (HPC) cluster operated by the Laboratory Computing Resource Center (LCRC) at Argonne National Laboratory. It consists of 6 nodes, each operating 2X AMD EPYC 7742 64-Core Processors and 8x NVIDIA A100 GPUs with 320GB of GPU memory, 1TB of DDR4 memory and 14TB of local scratch. 

\subsection{Evaluated Applications}\label{subsec:apps}
We evaluate the \tech methodology by - (1) comparing it to existing  the state-of-the-art methods to design SNN based transformer models, and (2) measuring the performance and scalability of \tech to design large scale foundation models. 

\subsubsection{Large Language Model}
NeuTransformer is also used to design large language models. The following datasets and models are used in the evaluation of NeuTransformer.

Datasets:
\begin{itemize}
    \item Shakesphere: 40,000 lines of Shakespeare from a variety of Shakespeare's plays.
    \item Openwebtext: An open-source replication of the WebText dataset from OpenAI.
\end{itemize}

Models:
\begin{itemize}
    \item GPT-2, GPT-2-Medium, GPT-2-Large: are transformer models pre-trained on a very large corpus of the Openwebtext dataset. This means it was pre-trained on the raw texts only and is trained to guess the next word in a sentence. Inputs are sequences of continuous text of a certain length and the targets are the same sequence, shifted one token (word or piece of word) to the right.
\end{itemize}

\subsection{Evaluated Metrics}\label{subsec:metrics}

\begin{itemize}
\item Token-wise Accuracy: Char to char comparison of the generated characters of ANN vs the SNN for an identical input sequence. 
\item Cosine Similarity: Cosine Similarity is a metric used to determine the cosine of the angle between two non-zero vectors in a multi-dimensional space.
\item Perplexity: Perplexity measures the quality of language models. It is calculated as exponent of the loss obtained from the model.
\item Bit-per-Byte: Bits-per-byte (bpb) measures the average number of bits required to predict the next token in a sequence.
\end{itemize}

\begin{table*}[t!]
	\renewcommand{\arraystretch}{1.2}
	\setlength{\tabcolsep}{2pt}
	\centering
	{\fontsize{8}{10}\selectfont
	\begin{tabular}{|l||c|c|c|c|c|c|}
		\hline

		Model & Dataset & Precision & Params  & \textbf{$E_{ASA}$} ($\mu$J) & \textbf{$E_{SSA}$} ($\mu$J)  
    		& $Throughput_{ANN}$ $(char/sec)$ \\
		\hline
		\hline
		
        \multirow{2}{*}{GPT-2-Small} &	\multirow{6}{*}{OpenWebText}	&	$FP_{16}$	&	\multirow{2}{*}{117}	&	48.93	&	57.11 & 3193.07	\\
    	&	&	$FP_{8}$	&		&	31.20	& -	 &  3831.11 \\
        \cline{1-1}\cline{3-7}
        \multirow{2}{*}{GPT2-Medium}	&		&	$FP_{16}$	&	\multirow{2}{*}{345}	&	195.7	& 221.7	& 623.30  	\\
        	&		&	$FP_{8}$	&		&	163.02	& -	& 714.31	\\
        \cline{1-1}\cline{3-7}
        \multirow{2}{*}{GPT-2-Large}	&		&	$FP_{16}$	&	\multirow{2}{*}{763}	&	382.01	& 514.30	 & 418.71 	\\
       	&		&	$FP_{8}$	&		& 449.61	& -  & 	549.26	\\
		\hline
	\end{tabular}}
\caption{Comparison of the performance of the SNN w.r.t to baseline ANN Transformer on   Hardware for FP-16 and FP-8 quantization.}
\label{tab:results_conversion}
\end{table*}

\begin{table*}[!t]

	\renewcommand{\arraystretch}{1.1}
	\setlength{\tabcolsep}{2pt}
	\centering
	{\fontsize{8}{10}\selectfont
	\begin{tabular}{|c||c|c|c|c|c|c|c|c|c|}
    \hline
    
\multirow{2}{*}{Model}      & \multirow{2}{*}{Dataset}     & \multirow{2}{*}{IPUs} & \multirow{2}{*}{Sequence} & \multicolumn{3}{|c|}{ANN} & \multicolumn{3}{|c|}{SNN} \\ \cline{5-10}

        &  &  & Length & Latency & Throughput & Power Consumption & Latency & Throughput  & Power Consumption  \\ 
\hline
\multirow{6}{*}{GPT-2-Small}  & \multirow{12}{*}{OpenWebtext} & \multirow{3}{*}{1}    & 16              & 0.4076             & 2453.38                    & 46.93                                & 0.3640             & 2747.25                 & 93.8                \\
           &             &      & 64              & 0.3434             & 2912.05                    & 54.01                                & 0.3801             & 2630.88                 & 131.03                             \\
           &             &      & 256             & 0.3779             & 2646.20                    & 83.16                                & 0.3814             & 3121.91                 & 183.29                                    \\
           \cline{3-10}
           &             & \multirow{3}{*}{2}    & 16              & 0.3808             & 2626.05                    & N/A                                  & 0.3424             & 2920.56                 & N/A                                   \\
           &             &      & 64              & 0.3587             & 2787.84                    & N/A                                  & 0.3783             & 2643.40                 & N/A                                        \\
           &             &      & 256             & 0.3794             & 2635.74                    & N/A                                  & 0.3894             & 2168.05                 & N/A                                 \\ \cline{1-1} \cline{3-10}
\multirow{6}{*}{GPT-2-Medium} &             & \multirow{3}{*}{4}    & 16              & 1.2902             & 775.07                     & 194.21                               & 1.0910             & 916.59                  & 224.91                                   \\
           &             &      & 64              & 1.1416             & 875.96                     & 228.31                               & 0.9491             & 1053.62                 & 283.13                                  \\
           &             &      & 256             & 1.1204             & 892.53                     & 230.54                               & 1.0834             & 823.02                  & 314.75                              \\\cline{3-10}
           &             & \multirow{3}{*}{8}    & 16              & 1.2062             & 829.04                     & N/A                                  & 1.0183             & 982.02                  & N/A                                      \\
           &             &      & 64              & 1.1560             & 865.05                     & N/A                                  & 0.9011             & 1109.75                 & N/A                               \\
           &             &      & 256             & 1.3034             & 767.22                     & N/A                                  & 1.2460             & 802.56                  & N/A                                   \\ 
           \hline
	\end{tabular}}
\caption{Comparison of the performance of the SNN w.r.t to baseline transformer on Graphcore hardware}
\label{tab:results_hardware}
\end{table*}

\subsection{Energy Evaluation} \label{subsec:energy}
We perform the energy evaluation of the ASA and SSA blocks by calculating the number of operations performed for a single input (I). The energy consumed to perform the operation are estimated using prior explorations \cite{li2024spikeformer,energyestimation2014Mark}. The two key operations performed in the ASA and SSA are the multiply-and-accumulate (MAC) and accumulate (AC). The energy consumption for these operations when estimated on a 45nm node \cite{energyestimation2014Mark} is $E_{MAC}$ = 4.6pJ and $E_{AC}$ = 0.9pJ, respectively.

\begin{equation}
	\label{eq:enrg_asa}
	E_{ASA} = E_{MAC} \times FLOPS(ASA(QKV))
\end{equation}

\begin{equation}
	\label{eq:enrg_ssa}
	E_{SSA} = E_{AC} \times SpikeOP(SSA)
\end{equation}

where, SpikeOP are the total number of spiking operations performed in the SSA block. An approximation of the number of SpikeOPs can be defined as follows: 

\begin{equation}
	\label{eq:enrg_}
	SpikeOP(SSA) = \sum{}{} I \times T \times S_{AvgRate} \times S_{OPs}
\end{equation}

where, I is the input to the SSA block, T is the time window of the spike input, $S_{AvgRate}$ is the average spike rate for inputs in a batch and $S_{OPs}$ is the total number of binary spiking operations performed in a single iteration of the SSA block.

\subsection{Throughput Evaluation of Graphcore Platfrom}
We evaluate the performance of the benchmark applications on Nvidia A100 GPUs and the Graphcore platform. Graphcore IPUs are designed to facilitate deep learning workloads by processing fine-grained operations across a large number of parallel threads. The ability to process individual threads on sub-blocks offers a two-fold benefit on SNN workloads over single-instruction-multiple-data/thread (SIMD/SIMT) GPUs: i) instructions from different network layers can be concurrently processed, where the constraints of contiguous vectorized data is no longer a performance bottleneck, and ii) MIMD processing can accelerate applications with irregular and sparse data access without incurring performance degradation. This is optimal for spike-based workloads which include additional processing overhead in computing the state-driven dynamics of spiking neuron models

\section{Results and Discussions}\label{sec:results}

We report the performance of the convert SNN versus its respective baseline in Table \ref{tab:results_conversion}. The baseline model and its equivalent SNN are evaluated for the metrics described in Section \ref{subsec:metrics}.

\subsection{Performance and Scalability on Language Models}

Comparison between the converted SNN versus its respective baseline across several key metrics is shown in Table \ref{tab:results_conversion}. Firstly, token-wise accuracy was assessed by comparing the output from both models given the same input sequence. The SNN showed character accuracy rates of 88.3\%, 84.9\%, 75.4\%, and 71.8\%, respectively, indicating a variance from the baseline with an increase in model size. Secondly, we utilized cosine similarity to measure the alignment of character vectors produced for a single batch of the test set. Notably, the cosine similarity decreased from 0.94 to 0.73 as the size of the model is scaled up. Thirdly, perplexity, a metric for evaluating language model quality, demonstrated a 9.1\% reduction in the SNN compared to the baseline. Finally, the bit-per-byte (bpb) metric, which indicates the average number of bits required to predict the next token, remained consistent across both models due to the fixed hyperparameter of the spike window encoding the character embeddings. The performance of the \tech model when scaling the size of the model can be further studied in Figure \ref{fig:plots}. Larger models, such as GPT-2 Medium and Large, do not demonstrate the expected performance for the measured metrics. This reduction in performance can attributed to the imperfect transformation of the feed-forward block into the spiking domain and limitations in the surrogate gradient-based learning used to fine-tune the SSA block.

\subsection{Analysis of Energy Estimation}
Table \ref{tab:results_conversion} shows the estimated energy consumption for all operations in the ASA block ($E_{ASA}$) and SSA block ($E_{SSA}$) for a single input (I). The methodology used to estimate the energy consumption is detailed in Section \ref{subsec:energy}. We observe that the estimated energy consumed to compute the SSA block reduces by 85.28\%, 85.22\%, 71.77\% and 64.71\%, respectively, when compared to the ASA block. To simplify this experiment, we assume that the ASA and SSA block are executed on a digital implementation of a multiply and accumulator (MAC). We expect the energy efficiency performance of the SSA to be further improved when implemented on a NmC. We also observe that the energy consumption performance of the SSA block deteriorates with an increase in the size of the language model. This can be attributed to the increase in the time window (T) of the input (I), as shown in equation \ref{eq:enrg_}, for larger implementations of the GPT-2 models. 

\subsection{Throughput Evaluation}
Table \ref{tab:results_hardware} compares the performance of the baseline vs its equivalent SNN when executed the Graphcore platform in terms of throughput, latency and power consumption. We also scale the baseline model in terms of sequence length to demonstrate the improved performance of the SNN in terms of application throughput (generated tokens per second). 

\section{Conclusions}\label{sec:conclusions}
In this work, we propose a methodology to design SNN-based language models for inference tasks, demonstrating a reduction in throughput and estimated energy consumption when deployed on neuromorphic hardware. Our methodology exploits both prior methods to design large-scale SNN and supervised fine-tuning to design transformer-based SNN. The proposed methodology works in two steps: 1) replacing self-attention (SA) mechanism with a SNN-based self-attention (SSA), and (2) fine-tuning the SSA block using SNN-based surrogate learning algorithms. We observe that the \tech methodology outperforms the state-of-the-art methods to train SNN-based transformers for a vision transformer benchmark. We also demonstrate the scalability of the proposed methodology on GPT-2 model of increasing model size. However, we also demonstrate the limits of the proposed methodology as we observe that for model sizes greater than 300M parameters, the performance of the converted SNN degrades beyond an acceptable threshold. We also observe the computational efficiency of the SNN-based transformer models by demonstrating between 64.71\% and 85.28\% reduction in estimated energy consumption when implementing the self-attention mechanism.

\bibliographystyle{neurips}
\bibliography{neurips_2025}

\end{document}